\documentclass[runningheads]{llncs}
\usepackage[T1]{fontenc}
\usepackage{graphicx}
\usepackage{float}
\floatstyle{ruled}
\newfloat{program}{t}{lop}
\floatname{program}{Program}
\usepackage{csquotes}

\newcommand{\etal}{\textit{et al.}}
\usepackage{multicol, latexsym}
\usepackage{blindtext}
\usepackage{subfigure}
\usepackage{caption}
\usepackage{graphicx}
\usepackage{calc}
\usepackage{ifthen}
\usepackage{algorithm}
\usepackage{algorithmic}
\usepackage{pgfplots}
\usepackage{wrapfig}
\usepackage{lipsum}

\begin{document}
\title{Genetic Algorithm for Program Synthesis}
%
%
\author{Yutaka Nagashima\inst{1}\orcidID{0000-0001-6693-5325}}
\authorrunning{Y. Nagashima}
%
\institute{
Independent, Cambridge, the UK\\
}
\maketitle              
\begin{abstract}
A deductive program synthesis tool takes a specification as input and 
derives a program that satisfies the specification.
The drawback of this approach is that search spaces for such correct programs tend to be enormous,
making it difficult to derive correct programs within a realistic timeout.
To speed up such program derivation,
we improve the search strategy of a deductive program synthesis tool, SuSLik, 
using evolutionary computation.
Our cross-validation shows that
the improvement brought by evolutionary computation
generalises to unforeseen problems.
\end{abstract}
\section{Introduction}
\label{sect:introduction}
A far-fetched goal of artificial intelligence research is to build a system
that writes computer programs for humans.
To achieve this goal, 
researchers take two distinct approaches:
deductive program synthesis and inductive program synthesis.
Both approaches attempt to produce programs requested by human users.
The difference lies how they produce programs: 
deductive synthesis tries to \textit{deduce} programs that satisfy specifications,
while inductive program synthesis tries to \textit{induce} programs from examples.

While such inductive synthesis alleviates the burden of implementation
by guessing programs from given input-output examples, in inductive synthesis resulting programs are not trustworthy. 
Deductive synthesis overcomes this limitation with formal specifications: 
it allows users to formalise \textit{what} they want as specifications, whereas inductive synthesis tools guess
\textit{how} programs should behave from examples provided by users.
Thus, in deductive synthesis providing formal specifications remains as users' responsibility.
The upside of deductive synthesis is, however, users can obtain \textit{correct} programs upon success.

SuSLik \cite{suslik_popl}, for example, is one of such deductive synthesis tools.
It takes a specification provided by humans and attempts to produce heap-manipulating programs satisfying the specification in a language that resembles the C language.
Internally, this derivation process is formulated as proof search:
SuSLik composes a heap-manipulating program by conducting a best-first search for a proof goal presented as specification.
The drawback is that the search algorithm often fails to find a proof within a realistic timeout.
That is, even we pass a specification to SuSLik,
SuSLik may not produce a program satisfying the specification.
According to Itzhaky \etal{} \cite{suslik_cav21}, \textit{different synthesis tasks benefit from different search parameters, and that we might need a mechanism to tune SuSLik ’s search strategy for a given synthesis task.}




\section{SuSLik's Search Strategy}\label{sec:susliks_search}

SuSLik synthesises a program by searching for a corresponding proof.
We can see SuSLik's proof search as an exploration of an OR-tree,
nodes of the tree represent (intermediate) synthesis goals,
while edges of the tree represent rule applications.
The shape of such search tree is not known in advance,
and the task of SuSLik is to identify a solved node,
in which a proof is complete.

\begin{wrapfigure}{r}{0.6\textwidth}
\vspace{-30pt}
    \begin{center}
    \includegraphics[width=1.0\linewidth]{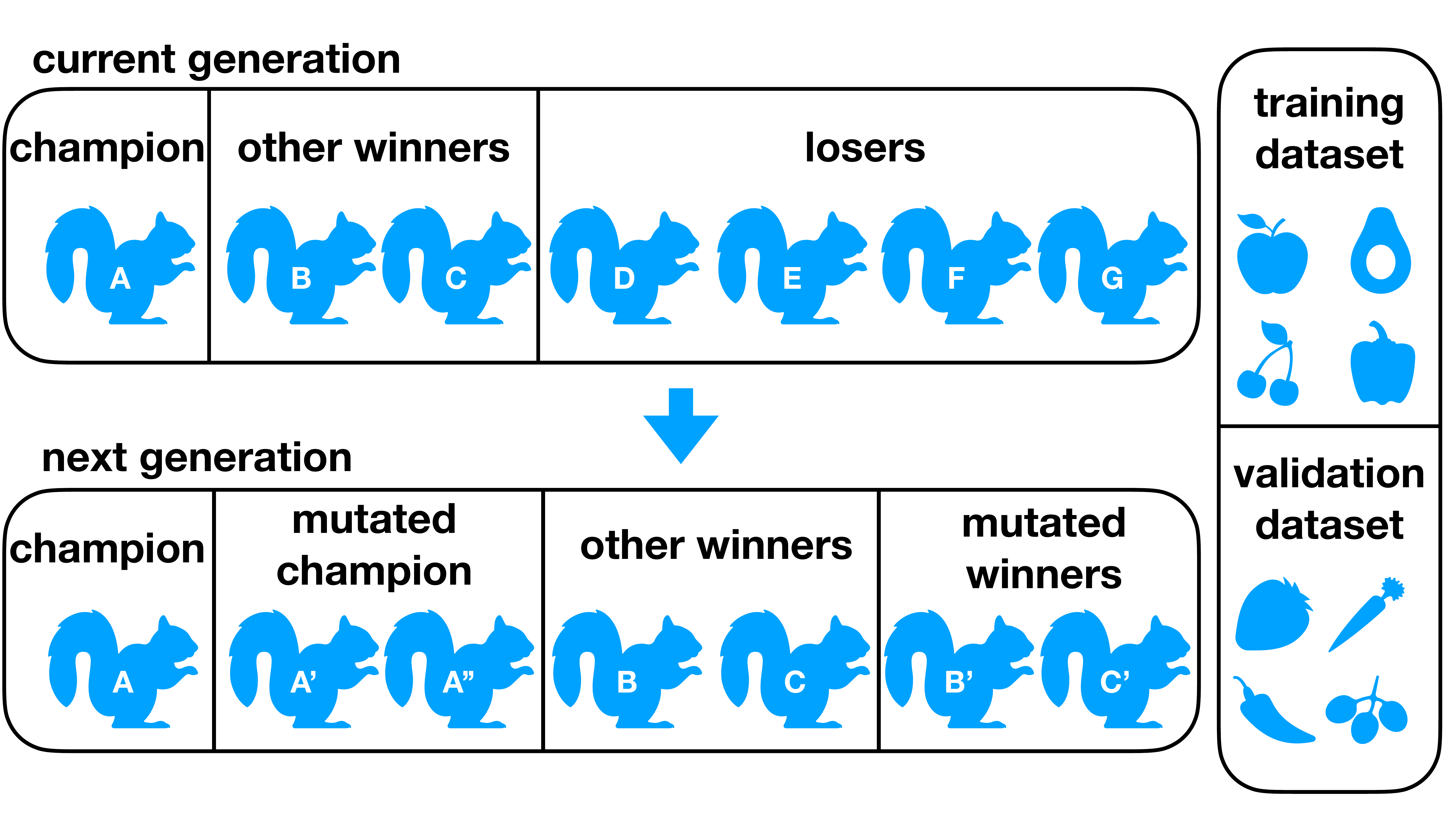}
    \end{center}
    \vspace{-20pt}
    \caption{Mutation and Elitist Selection}
    \vspace{-20pt}
    \label{fig:evolution}
\end{wrapfigure}

Since such OR-trees can be too large to find proofs within a realistic timeout,
SuSLik narrows the search space using a \textit{proof strategy}.
Essentially, proof strategy in SuSLik is a function
that takes a synthesis goal and returns an ordered list of rules to apply next.
Itzhaky \etal{} developed the default strategy
by manually encoding human expertise.
For example, the default strategy precludes the application of a rule called \texttt{CALL}
when another rule \texttt{CLOSE} has been applied before reaching the current node.
This way, the SuSLik rules are grouped into 10 ordered lists,
and the order of these rules in the lists define 
how SuSLik explores the corresponding OR-tree.

Another decision SuSLik has to make for an effective search
is to select the next node to expand.
The current version of SuSLik make this decision using a cost function,
manually developed and tuned by Itzhaky \etal{} \cite{suslik_cav21}.

Both the weights of the cost function and orders of derivation rules are
manually tuned for the benchmark used in their evaluation \cite{suslik_cav21};
however, as we show in Section \ref{sec:evaluation},
our evolutionary framework finds better strategies through evolution.
\section{Evolutionary Computation for SuSLik}\label{sec:architeture}

The aim of our evolutionary computation is to optimise 
the order of each group of derivation rules
and the weights of the cost function, which is used to implement best-first search.

Algorithm \ref{alg:genetic_algorithm} summarises the genetic algorithm
we used in our framework to improve the search strategy of SuSLik.
Firstly, the algorithm takes a set of training problems an inputs,
using which we evolve SuSLik instances over 40 generations.
Line 1 defines the initial population.
Each individual in a population is evaluated according to the fitness function described in Section \ref{sec:fitness}.

For each generation,
we copy individuals from the previous iteration (Line 6),
mutate them (Line 7),
evaluate individuals (Line 8).
Then, we sort all individuals in the current generation based on their performance (Line 9 - 10).
And we continue to the next generation using
the best 20 individuals from the current pool.
In the following, we explain the mutation algorithm,
the fitness function,
and our selection algorithm.

\paragraph{Mutation.}\label{sec:mutation}
As we explained in Section \ref{sec:susliks_search},
by default a search strategy of SuSLik is defined by two factors:
the order of rule application and weights of 
each node in the search tree.
To determine an effective way to apply genetic algorithms to program synthesis in SuSLik,
we implemented the following three different mutation algorithms:
\begin{itemize}
    \item \textit{Order-only mutation} changes only the order of rule application for each node.
    \item \textit{General rule-weight mutation} changes the weights of each node based on what rules have been applied to reach that node.
    \item \textit{Goal-specific rule-weight mutation} 
    allows SuSLik to choose a weight for each rule based on properties of a node during a search.
\end{itemize}

\begin{wrapfigure}{R}{0.5\textwidth}
\vspace{-40pt}
\begin{minipage}{0.5\textwidth}
\begin{algorithm}[H]
\caption{Evolutionary Computation for SuSLik}
\label{alg:algorithm}
\begin{flushleft}
\textbf{Input}: synthesis problems for SuSLik \\
\textbf{Output}: a SuSLik search strategy
\end{flushleft}
\begin{algorithmic}[1] 
\STATE Let $old\_pop$ be the initial population.
\STATE fitness($old\_pop$)
\STATE $generation \gets 1$
\WHILE{$generation$ $\le$ $40$}
\STATE $generation \gets generation + 1$
\STATE $new\_pop \gets old\_pop$
\STATE mutate($new\_pop$)
\STATE fitness($new\_pop$)
\STATE $whole\_pop \gets old\_pop + new\_pop$ 
\STATE sort ($whole\_pop$)
\STATE $old\_pop \gets$ take ($whole\_pop$, 20)
\ENDWHILE
\end{algorithmic}
\label{alg:genetic_algorithm}
\end{algorithm}
\vspace{-40pt}
\end{minipage}
\end{wrapfigure}

\paragraph{Fitness.}\label{sec:fitness}

The fitness function measures the performance of SuSLik instances.
More specifically, it measures how many derivation problems each SuSLik instance solves
within the timeout of 2.5 seconds for each problem. 
When multiple SuSLik instances solve the same number of derivation problems, 
the fitness function uses the numbers of rules fired by the instances as a tie-breaker:
it considers that the instance that solves a certain number of problems with a smaller number of rule applications is better than another instance that solves the same number of problems with a larger number of rule applications.

\paragraph{Selection.}\label{sec:selection}
We adopt a version of \textit{elitist selection} as our selection method:
we pass individuals from the current generation to the next generation.
by copying them and mutating them
if they show better performance in the current generation.
Figure \ref{fig:evolution} provides the schematic view of our elitist selection.
Unlike the standard elitist selection algorithm,
ours prioritizes the best individual in each generation to speed up the evolution:
the best individual in each generation, called \textit{champion},
is entitled with three children,
one original copy without mutation and two mutated children,
whereas each of other 19 winners has one original copy
and only one mutated child in the next generation.

\paragraph{}Note that each individual has two kinds of properties to mutate:
the order of derivation rules, and weights used in the cost function.
While we represent the weights as floating point numbers,
we adopt permutation encoding for the orders of derivation rules.

For each permutation encoding, 
each individual has the probability of 0.1 to be moved,
while we change weights by multiplying a random number between 0.8 and 1.2.
In our framework, we do not apply crossover to permutation encoding:
since our sequences denoting rule orders tend to be short,
we are not sure if crossovers would result in a better performance of evolution.

Our evolutionary computation for program synthesis differs from genetic programming \cite{genetic_programming_koza} or 
evolutionary programming \cite{evolutionary_programming_fogel}:
we did not directly apply simulated evolution to programs,
but our framework improves the search mechanism for deriving correct programs through evolution.
We take this approach to take the best of both worlds:
the correctness of resulting programs guaranteed by the deductive synthesis
and its certification tool,
and the search heuristics enhanced through evolutionary computation.

\section{Evaluation} \label{sec:evaluation}

We conducted cross-validations to evaluate
what improvements our evolutionary computation framework brought to SuSLik.
We measured how many synthesis problems
SuSLik failed to solve with in 2.5 seconds of timeout.
For this evaluation,
we used a consumer laptop running Ubuntu 20.04.3 LTS
on a machine with 16 CPUs of 
AMD Ryzen 7 4,800H with Radeon Graphics
and 15,854MB of main memory.

As SuSLik is a new tool,
we have only 65 problems available in our benchmark:
problems from a preceding work on SuSLik \cite{suslik_cav21} 
and new problems prepared for this project.
These problems include tasks on various data-structures such as
integers, singly linked lists, sorted lists, doubly linked lists, lists of lists, binary trees, and packed trees.

Firstly, we randomly split our benchmarks into two groups:
the validation dataset and training dataset.
Then, using the training dataset 
we apply our evolutionary computation described in Algorithm \ref{alg:genetic_algorithm}
to evolve SuSLik's search strategy.
As explained in Section \ref{sec:architeture}, the output of our evolutionary computation is 
just one search strategy produced after 40 generations.
However, in this experiment we conducted cross-validations using the best individual 
from the training set for each generation to see how our framework produces transferable improvement over generations.

To reduce the influence from a specific random split,
we conducted this experiment four times,
and the result of each experiment is illustrated from Figure 2 to Figure 5.
In these figures,
the horizontal axes represent the number of generations,
while the vertical axes represent 
the number of synthesis problems SuSLik did \textit{not} solve within the timeout.

These figures show that when adopting the general rule-weight mutation, 
our evolutionary framework managed to improve SuSLik's capability to find solutions in validation sets, even though evolution is based on training sets.
That is, somewhat contrarily to the prediction by Itzhaky introduced in Section \ref{sect:introduction}, we found that
there are strategies that tend to perform better for unforeseen problems,
and we can find such strategies using evolutionary computation.

On the other hand, the order-only mutation and goal-specific rule-weight mutation resulted in less promising results.
In particular, the goal-specific rule-weight mutation over-fitted to training data in Figure 2 and Figure 5, probably due to its capability to fine tune the strategy for our small dataset.

\begin{figure}[tb]
\centering
\begin{minipage}{0.5\textwidth}
\centering
\resizebox{\textwidth}{!}{%
\begin{tikzpicture}{scale=0.5}
\tikzstyle{every node}=[font=\large]
\begin{axis}[
    xlabel={Generation.},
    ylabel={Unsolved problems out of 33.},
    xmin=0, xmax=42,
    ymin=15, ymax=23, 
    xtick={0,10,20,30,40},
    ytick={0, 5, 10,12,14,16,18,20,22,24, 30},
    legend pos=north east,
    grid style=dashed,
    only marks
]

\addplot[
    color=black,
    mark=triangle,
    ]
    coordinates {
    (0, 20)  (10, 19) (20, 19)  (30, 19) (40, 18)
    };
    ];
    \addlegendentry{order-only};

\addplot[
    color=black,
    mark=square,
    ]
    coordinates {
    (0, 22)  (10, 18) (20, 16)  (30, 17) (40, 16)
    };
    ];
    \addlegendentry{general rule-weight};
    
\addplot[
    color=black,
    mark=star,
    ]
    coordinates {
    (0, 20)  (10, 20) (20, 20)  (30, 20) (40, 19)
    };
    ];
    \addlegendentry{goal-specific rule-weight};
    
\end{axis}

\end{tikzpicture}
}
\caption{Cross-validation 1}
\label{fig:experiment1}
\end{minipage}%
\begin{minipage}{0.5\textwidth}
\centering
\resizebox{\textwidth}{!}{%
\begin{tikzpicture}{scale=0.5}
\tikzstyle{every node}=[font=\large]
\begin{axis}[
    xlabel={Generation.},
    ylabel={Unsolved problems out of 24.},
    xmin=0, xmax=42,
    ymin=13, ymax=18.5, 
    xtick={0,10,20,30,40},
    ytick={0, 5, 10,12,14,16,18,20,22,24, 30},
    legend pos=south west,
    grid style=dashed,
    only marks
]

\addplot[
    color=black,
    mark=triangle,
    ]
    coordinates {
    (0, 18)  (10, 17) (20, 16)  (30, 16) (40, 15)
    };
    ];
    \addlegendentry{order-only};

\addplot[
    color=black,
    mark=square,
    ]
    coordinates {
    (0, 17)  (10, 16) (20, 16)  (30, 15) (40, 15)
    };
    ];
    \addlegendentry{general rule-weight};
    
\addplot[
    color=black,
    mark=star,
    ]
    coordinates {
    (0, 18)  (10, 18) (20, 18)  (30, 17) (40, 18)
    };
    ];
    \addlegendentry{goal-specific rule-weight};
    
\end{axis}
\end{tikzpicture}
}
\caption{Cross-validation 2}
\label{fig:experiment2}
\end{minipage}
\end{figure}
\begin{figure}[tb]
\centering
\begin{minipage}{0.5\textwidth}
\centering
\resizebox{\textwidth}{!}{%
\begin{tikzpicture}
\tikzstyle{every node}=[font=\large]
\begin{axis}[
    xlabel={Generation.},
    ylabel={Unsolved problems out of 34.},
    xmin=0, xmax=42,
    ymin=13, ymax=22, 
    xtick={0,10,20,30,40},
    ytick={0, 5, 10,12,14,16,18,20,22,24, 30},
    legend pos=south west,
    grid style=dashed,
    only marks
]

\addplot[
    color=black,
    mark=triangle,
    ]
    coordinates {
    (0, 21)  (10, 21) (20, 21)  (30, 21) (40, 20)
    };
    ];
    \addlegendentry{order-only};

\addplot[
    color=black,
    mark=square,
    ]
    coordinates {
    (0, 22)  (10, 18) (20, 18)  (30, 16) (40, 16)
    };
    ];
    \addlegendentry{general rule-weight};
    
\addplot[
    color=black,
    mark=star,
    ]
    coordinates {
    (0, 21)  (10, 20) (20, 21)  (30, 20) (40, 20)
    };
    ];
    \addlegendentry{goal-specific rule-weight};
    
\end{axis}
\end{tikzpicture}
}
\caption{Cross-validation 3}
\label{fig:experiment3}
\end{minipage}%
\begin{minipage}{0.5\textwidth}
\centering
\resizebox{\textwidth}{!}{%
\begin{tikzpicture}
\tikzstyle{every node}=[font=\large]
\begin{axis}[
    xlabel={Generation.},
    ylabel={Unsolved problems out of 34.},
    xmin=0, xmax=42,
    ymin=14, ymax=25, 
    xtick={0,10,20,30,40},
    ytick={0, 5, 10,12,14,16,18,20,22,24, 30},
    legend pos=south west,
    grid style=dashed,
    only marks
]

\addplot[
    color=black,
    mark=triangle,
    ]
    coordinates {
    (0, 24)  (10, 23) (20, 23)  (30, 23) (40, 22)
    };
    ];
    \addlegendentry{order-only};

\addplot[
    color=black,
    mark=square,
    ]
    coordinates {
    (0, 24)  (10, 21) (20, 21)  (30, 18) (40, 18)
    };
    ];
    \addlegendentry{general rule-weight};
    
\addplot[
    color=black,
    mark=star,
    ]
    coordinates {
    (0, 23)  (10, 23) (20, 22)  (30, 22) (40, 24)
    };
    ];
    \addlegendentry{goal-specific rule-weight};
    
\end{axis}

\end{tikzpicture}
}
\caption{Cross-validation 4}
\label{fig:experiment4}
\end{minipage}
\end{figure}

\section{Discussion}

The limited size of available dataset is the main challenge
we faced in this project.
This problem is partially unavoidable since program synthesis itself is still an emerging field in Computer Science.
Other AI projects for interactive theorem provers take advantage of
large existing proof corpora for training.
For example, Nagashima built a tactic prediction tool, PaMpeR \cite{pamper}, for Isabelle/HOL 
by extracting 425,334 data points \cite{dataset} from the Archive of Formal Proofs (AFP) \cite{afp}.
Li \etal{} also mined the AFP and produced 820K training examples for conjecturing.
For Coq, Yang \etal{} constructed a dataset containing 71K proofs
from 123 projects \cite{coqgym}, 
whereas Huang \etal{} \cite{gamepad} extracted a dataset consisting of 1,602 lemmas from
the Feit-Thompson formalization.
For HOL Light \cite{holight},
The HOLStep \cite{holstep} used 1,013,046 training examples and 196,030 testing examples extracted from 11,400 proofs,
while the HOList project presented a benchmark based on 2,199 definitions 
and 29,462 theorems and lemmas.
These projects managed to gather large data sets since their underlying theorem provers,
Isabelle/HOL, Coq, and HOL Light, have a larger user base than SuSLik \cite{suslik_popl} does.

For the moment, 
our framework improves \textit{static} parameters for SuSLik.
That is, the resulting weights and rule orders are fixed for all intermediate synthesis problems.
Our evaluation has shown that our static parameter optimisation (general rule-weight mutation) using evolutionary computation
generalises well:
a SuSLik instance that performs well for a training dataset
tends to perform well for an evaluation dataset.
We expected that we could achieve even better performance
by producing \textit{dynamic} parameters (goal-specific rule-weight) for SuSLik:
functions that inspect a node at hand and decide on a promising rule order and weights for that node.
Our efforts in this direction are, unfortunately, unsuccessful so far.
We hope that a larger training dataset would allow for such optimisation in the future.

\section{Related Work} \label{sec:related_work}




Despite the current trend of applying deep learning to theorem proving,
we consciously avoided deep learning in this project,
as we have a limited number of synthesis problems at hand.

Even though there was an attempt to use reinforcement learning \cite{reinforcement_learning} for a connection-style proof search 
\cite{reinforcement_learning_of_TP};
we mindfully chose evolutionary computation over reinforcement learning:
since we do not have a changing environment in our setting,
it is unclear if we gain any benefits from having two metrics, 
reward function for the long term goal and
value function for the short term benefit.
Instead, we improved SuSLik's default search strategy
for randomly chosen fixed training problem sets
and measured how the improvement generalizes to validation sets.

When implementing our framework for evolutionary computation,
we took the advantage of a Python framework for evolutionary computation called DEAP \cite{deap},
even though SuSLik itself is implemented in Scala.

Previously, we attempted to improve proof strategies \cite{psl} for Isabelle/HOL using evolutionary computing \cite{evolutionary_isabelle}.
However, the focus of that project shifted to the prediction of induction arguments \cite{smart_induction,faster_smarter_induction} using meta-languages \cite{lifter,selfie}.

Nawaz \etal{} used a genetic algorithm to evolve random proof sequences
to target proofs.
The drawback of their approach is that the fitness function used in the genetic algorithm
relies on the existence of a proof for a given problem.
Therefore, this framework is not applicable to open conjectures without existing proofs
\cite{genetic_hol4}.




\section*{Acknowledgment}
We would like to thank Andreea Costea for preparing additional SuSLik problems for cross-validations.

%
%
%
\bibliographystyle{splncs04}
\bibliography{main}

\begin{thebibliography}{10}
\providecommand{\url}[1]{\texttt{#1}}
\providecommand{\urlprefix}{URL }
\providecommand{\doi}[1]{https://doi.org/#1}

\bibitem{evolutionary_programming_fogel}
Fogel, L., Owens, A.J., Walsh, M.J.: Artificial intelligence through simulated
  evolution (1966)

\bibitem{deap}
Fortin, F.A., {De Rainville}, F.M., Gardner, M.A., Parizeau, M., Gagn\'e, C.:
  {DEAP}: Evolutionary algorithms made easy. Journal of Machine Learning
  Research  \textbf{13},  2171--2175 (jul 2012)

\bibitem{holight}
Harrison, J.: {HOL} {L}ight: {A} tutorial introduction. In: Srivas, M.K.,
  Camilleri, A.J. (eds.) Formal Methods in Computer-Aided Design, First
  International Conference, {FMCAD} '96, Palo Alto, California, USA, November
  6-8, 1996, Proceedings. Lecture Notes in Computer Science, vol.~1166, pp.
  265--269. Springer (1996). \doi{10.1007/BFb0031814},
  \url{https://doi.org/10.1007/BFb0031814}

\bibitem{gamepad}
Huang, D., Dhariwal, P., Song, D., Sutskever, I.: Gamepad: {A} learning
  environment for theorem proving. In: 7th International Conference on Learning
  Representations, {ICLR} 2019, New Orleans, LA, USA, May 6-9, 2019.
  OpenReview.net (2019), \url{https://openreview.net/forum?id=r1xwKoR9Y7}

\bibitem{suslik_cav21}
Itzhaky, S., Peleg, H., Polikarpova, N., Rowe, R.N.S., Sergey, I.: Deductive
  synthesis of programs with pointers: Techniques, challenges, opportunities -
  (invited paper). In: Silva, A., Leino, K.R.M. (eds.) Computer Aided
  Verification - 33rd International Conference, {CAV} 2021, Virtual Event, July
  20-23, 2021, Proceedings, Part {I}. Lecture Notes in Computer Science, vol.
  12759, pp. 110--134. Springer (2021). \doi{10.1007/978-3-030-81685-8\_5},
  \url{https://doi.org/10.1007/978-3-030-81685-8\_5}

\bibitem{holstep}
Kaliszyk, C., Chollet, F., Szegedy, C.: Holstep: {A} machine learning dataset
  for higher-order logic theorem proving. In: 5th International Conference on
  Learning Representations, {ICLR} 2017, Toulon, France, April 24-26, 2017,
  Conference Track Proceedings. OpenReview.net (2017),
  \url{https://openreview.net/forum?id=ryuxYmvel}

\bibitem{reinforcement_learning_of_TP}
Kaliszyk, C., Urban, J., Michalewski, H., Ols{\'{a}}k, M.: Reinforcement
  learning of theorem proving. In: Bengio, S., Wallach, H.M., Larochelle, H.,
  Grauman, K., Cesa{-}Bianchi, N., Garnett, R. (eds.) Advances in Neural
  Information Processing Systems 31: Annual Conference on Neural Information
  Processing Systems 2018, NeurIPS 2018, December 3-8, 2018, Montr{\'{e}}al,
  Canada. pp. 8836--8847 (2018),
  \url{https://proceedings.neurips.cc/paper/2018/hash/55acf8539596d25624059980986aaa78-Abstract.html}

\bibitem{afp}
Klein, G., Nipkow, T., Paulson, L., Thiemann, R.: The Archive of Formal Proofs
  (2004), \url{https://www.isa-afp.org/}

\bibitem{genetic_programming_koza}
Koza, J.R.: Genetic programming - on the programming of computers by means of
  natural selection. Complex adaptive systems, {MIT} Press (1993)

\bibitem{lifter}
Nagashima, Y.: Li{F}t{E}r: Language to encode induction heuristics for
  {I}sabelle/{HOL}. In: Lin, A.W. (ed.) Programming Languages and Systems -
  17th Asian Symposium, {APLAS} 2019, Nusa Dua, Bali, Indonesia, December 1-4,
  2019, Proceedings. Lecture Notes in Computer Science, vol. 11893, pp.
  266--287. Springer (2019). \doi{10.1007/978-3-030-34175-6\_14},
  \url{https://doi.org/10.1007/978-3-030-34175-6\_14}

\bibitem{evolutionary_isabelle}
Nagashima, Y.: Towards evolutionary theorem proving for {I}sabelle/{HOL}. In:
  L{\'{o}}pez{-}Ib{\'{a}}{\~{n}}ez, M., Auger, A., St{\"{u}}tzle, T. (eds.)
  Proceedings of the Genetic and Evolutionary Computation Conference Companion,
  {GECCO} 2019, Prague, Czech Republic, July 13-17, 2019. pp. 419--420. {ACM}
  (2019). \doi{10.1145/3319619.3321921},
  \url{https://doi.org/10.1145/3319619.3321921}

\bibitem{selfie}
Nagashima, Y.: Definitional quantifiers realise semantic reasoning for proof by
  induction. CoRR  \textbf{abs/2010.10296} (2020),
  \url{https://arxiv.org/abs/2010.10296}

\bibitem{dataset}
Nagashima, Y.: Simple dataset for proof method recommendation in isabelle/hol.
  In: Benzm{\"{u}}ller, C., Miller, B.R. (eds.) Intelligent Computer
  Mathematics - 13th International Conference, {CICM} 2020, Bertinoro, Italy,
  July 26-31, 2020, Proceedings. Lecture Notes in Computer Science, vol. 12236,
  pp. 297--302. Springer (2020). \doi{10.1007/978-3-030-53518-6\_21},
  \url{https://doi.org/10.1007/978-3-030-53518-6\_21}

\bibitem{smart_induction}
Nagashima, Y.: Smart induction for {I}sabelle/{HOL} (tool paper). In: 2020
  Formal Methods in Computer Aided Design, {FMCAD} 2020, Haifa, Israel,
  September 21-24, 2020. pp. 245--254. {IEEE} (2020).
  \doi{10.34727/2020/isbn.978-3-85448-042-6\_32},
  \url{https://doi.org/10.34727/2020/isbn.978-3-85448-042-6\_32}

\bibitem{faster_smarter_induction}
Nagashima, Y.: Faster smarter proof by induction in {I}sabelle/{HOL}. In: Zhou,
  Z. (ed.) Proceedings of the Thirtieth International Joint Conference on
  Artificial Intelligence, {IJCAI} 2021, Virtual Event / Montreal, Canada,
  19-27 August 2021. pp. 1981--1988. ijcai.org (2021).
  \doi{10.24963/ijcai.2021/273}, \url{https://doi.org/10.24963/ijcai.2021/273}

\bibitem{pamper}
Nagashima, Y., He, Y.: {P}a{M}pe{R}: proof method recommendation system for
  {I}sabelle/{HOL}. In: Huchard, M., K{\"{a}}stner, C., Fraser, G. (eds.)
  Proceedings of the 33rd {ACM/IEEE} International Conference on Automated
  Software Engineering, {ASE} 2018, Montpellier, France, September 3-7, 2018.
  pp. 362--372. {ACM} (2018). \doi{10.1145/3238147.3238210},
  \url{https://doi.org/10.1145/3238147.3238210}

\bibitem{psl}
Nagashima, Y., Kumar, R.: A proof strategy language and proof script generation
  for {I}sabelle/{HOL}. In: de~Moura, L. (ed.) Automated Deduction - {CADE} 26
  - 26th International Conference on Automated Deduction, Gothenburg, Sweden,
  August 6-11, 2017, Proceedings. Lecture Notes in Computer Science, vol.
  10395, pp. 528--545. Springer (2017). \doi{10.1007/978-3-319-63046-5\_32},
  \url{https://doi.org/10.1007/978-3-319-63046-5\_32}

\bibitem{genetic_hol4}
Nawaz, M.Z., Hasan, O., Nawaz, M.S., Fournier{-}Viger, P., Sun, M.: Proof
  searching in {HOL4} with genetic algorithm. In: Hung, C., Cern{\'{y}}, T.,
  Shin, D., Bechini, A. (eds.) {SAC} '20: The 35th {ACM/SIGAPP} Symposium on
  Applied Computing, online event, [Brno, Czech Republic], March 30 - April 3,
  2020. pp. 513--520. {ACM} (2020). \doi{10.1145/3341105.3373917},
  \url{https://doi.org/10.1145/3341105.3373917}

\bibitem{suslik_popl}
Polikarpova, N., Sergey, I.: Structuring the synthesis of heap-manipulating
  programs. Proc. {ACM} Program. Lang.  \textbf{3}({POPL}),  72:1--72:30
  (2019). \doi{10.1145/3290385}, \url{https://doi.org/10.1145/3290385}

\bibitem{reinforcement_learning}
Sutton, R.S., Barto, A.G.: Reinforcement learning: An introduction. {IEEE}
  Trans. Neural Networks  \textbf{9}(5),  1054--1054 (1998).
  \doi{10.1109/TNN.1998.712192}, \url{https://doi.org/10.1109/TNN.1998.712192}

\bibitem{coqgym}
Yang, K., Deng, J.: Learning to prove theorems via interacting with proof
  assistants. In: Chaudhuri, K., Salakhutdinov, R. (eds.) Proceedings of the
  36th International Conference on Machine Learning, {ICML} 2019, 9-15 June
  2019, Long Beach, California, {USA}. Proceedings of Machine Learning
  Research, vol.~97, pp. 6984--6994. {PMLR} (2019),
  \url{http://proceedings.mlr.press/v97/yang19a.html}

\end{thebibliography}

\end{document}